\documentclass[11pt,a4paper, final, twoside]{article}
\usepackage{amsmath}
\usepackage{fancyhdr}
\usepackage{amsthm}
\usepackage{amsfonts}
\usepackage{amssymb}
\usepackage{amscd}
\usepackage{latexsym}
\usepackage{amsthm}
\usepackage{graphicx}
\usepackage{graphics}
\usepackage{afterpage}
\usepackage[colorlinks=true, urlcolor=blue,  linkcolor=black, citecolor=black]{hyperref}
\usepackage{color}
\usepackage[table,xcdraw]{xcolor}
\usepackage{float}
\usepackage[linesnumbered,ruled,vlined, algo2e]{algorithm2e}

\usepackage{pdflscape}

\SetCommentSty{mycommfont}

\SetKwInput{KwInput}{Input}                
\SetKwInput{KwOutput}{Output}              

\theoremstyle{proposition}

\theoremstyle{definition}

\theoremstyle{remark}

\numberwithin{equation}{section}
\begin{document}
\hyphenpenalty=100000

\setlength{\arrayrulewidth}{0.4mm}
\setlength{\tabcolsep}{12pt}
\renewcommand{\arraystretch}{1.5}

\begin{flushright}

{\Large \textbf{\\A Meta-Heuristic Search Algorithm based on Infrasonic Mating Displays in Peafowls }}\\[5mm]
{\large \textbf{Kenekayoro Patrick$^\mathrm{*1}$\footnote{\emph{*Corresponding author: E-mail: Patrick.Kenekayoro@outlook.com}}  }}\\[1mm]
$^\mathrm{1}${\footnotesize \it Department of Mathematics and Computer Science, Niger Delta University, \\ Amassoma, Bayelsa State,
Nigeria.}
\end{flushright}

{\Large \textbf{Abstract}}\\[4mm]
\fbox{%
\begin{minipage}{5.4in}{\footnotesize  Meta-heuristic techniques are important
as they are used to find solutions to computationally intractable problems.
Simplistic methods such as exhaustive search become computationally 
expensive and unreliable as the solution space for search algorithms 
increase. As no method is guaranteed to perform better than all others in 
all classes of optimization search problems, there is a need to constantly 
find new and/or adapt old search algorithms. This research proposes an 
Infrasonic Search Algorithm, inspired from the Gravitational Search 
Algorithm and the mating behaviour in peafowls. The Infrasonic Search 
Algorithm identified competitive solutions to 23 benchmark unimodal and 
multimodal test functions compared to the Genetic Algorithm, Particle Swarm 
Optimization Algorithm and the Gravitational Search Algorithm. 
} \end{minipage}}\\[1mm]

\section{Introduction}\label{I1}
Researchers have shown the unsuitability of simplistic methods such as exhaustive search for solving optimization search problems. Simplistic methods particularly become increasingly inefficient as the solution space increases, and at times are unable to escape local minima.

Over the years, there have been increased interests in the adaptation of natural or biological phenomena to design more efficient optimization search algorithms. A particular class (population based) of these algorithms which have been shown to be efficient are based on cooperation among agents in a solution space. An agent can been seen as a candidate solution to the optimization problem to be solved. To find optimal solutions, candidate solutions share/exchange information about the search space as the algorithm progresses.

The main difference between these algorithms based on agent cooperation lies in the way the agents share information, and how candidate solutions are updated based on the exchanged information.

There is always a need to identify new or adapt existing meta-heuristic algorithms, which may perform better for some kind of computationally intractable problems. This is why this study proposes the Infrasonic Search Algorithm (ISA); a population based algorithm inspired from the Gravitational Search Algorithm (GSA)\cite{Rashedi2009}. 

In the GSA, it is assumed that masses represent the fitness of a solution, with heavier masses indicating fitter solutions and based on the law of gravitation, heavier masses attract lighter masses towards itself. In the same vein, in the infrasonic search algorithm higher quality solutions produce higher sound intensity compared to poorer solutions and agents move towards regions producing high intensity infrasonic sounds.

The difference between the GSA and the mating display algorithm is in the formulation of fitness of solutions and the movement of agents in a solution space. The GSA is based on the law of gravity, while the mating display heuristic is based on sound intensity. In subsequent sections, some population based heuristic algorithms are overviewed, the Infrasonic Search Algorithm is defined and result of solving 23 benchmark functions with the Infrasonic Search Algorithm compared with other population based algorithms is presented.

\section{Related Work}\label{I2}
It is necessary to find optimal combinations for the efficient management of scarce resources, which is why optimization methods are useful and have been applied in a number of fields among which include Manufacturing\cite{Alaouchiche2021,Yildiz2017},  Engineering\cite{Houssein2020,Feng2021} and Education\cite{kenekayoro2020population, Kenekayoro2016}.

Solutions to optimization problems which could be constrained or unconstrained, single objective or multi objective have been successfully found using 
algorithms that could broadly be classified into nature inspired vs non-nature inspired. However, Beheshti, Mariyam and Shamsuddin\cite{Beheshti2013} 
have listed other possible classification schemes from different view points. Irrespective of the classification scheme, algorithms can still be 
grouped to be either nature or non-nature inspired.

In this research, nature inspired algorithms are those algorithms such as the Genetic Algorithm\cite{Holland1975}, Ants Colony Optimization algorithm\cite{Dorigo1999}, Gravitational Search Algorithm\cite{Rashedi2009} and Particle Swarm Optimization Algorithm\cite{Venter2003} that mimic a physical or biological process that exist in nature, while non-nature inspired algorithms such as Tabu Search\cite{Glover1989}, Hill Climbing and 
Harmony Search\cite{Mahdavi2007} are based on heuristic not present in nature. 

Population based algorithms are a subset of nature inspired optimization methods where information is shared between candidate solutions in order to 
find better candidates. The difference between population based techniques is largely in the way information is shared between candidate solutions 
and how new candidate solutions are found based on the shared information. 

The Genetic Algorithm\cite{Holland1975}, inspired from natural evolution shares information by selection, crossover and mutation genetic operators which have been discussed extensively in researches\cite{Deb2000,MagalhaesMendes2013}. In the genetic algorithm, a candidate solution is represented as a chromosome 
and then two solutions are combined through the crossover operator or a single allele in a chromosome is changed through mutation to generate 
a new candidate solution. Selection techniques among which include tournament and roulette wheel are used to determine chromosomes that will make up a subsequent generation. Razali and Geraghty\cite{Razali2011} have investigated how different GA selection strategies influence the performance of the genetic algorithm. In principle the good properties of older solutions will be transferred to subsequent generations, ensuring that solutions found in subsequent generations are fitter than solutions found in previous generations. 

The Ants Colony Optimization Algorithm\cite{Dorigo1999} is inspired from the way ants find a path to a food source through cooperation. Initially, 
agents move randomly but update their path trail with pheromones proportional to the quality of a solution found. Subsequent agents use this pheromone 
trail as information that guide their traversal to a solution and then if the path is promising, additional pheromones are deposited on the path, thus 
increasing the likelihood that future agents will follow that path.

Agents in the Gravitational Search Algorithm\cite{Rashedi2009} share information following the Newtonian laws of gravity and motion. The quality of a solution (fitness of an agent) determines its mass, and agents move towards each other by the gravitational force $F = ma$. Fitter agents have larger masses, thus their movement is less influenced by less fit solutions, whilst unfit solutions move towards heavier masses (fitter solutions).

Agents in the Particle Swarm Optimization algorithm are guided to newer solutions based on the history of the agents' best solution found and the global best solution of all agents in a population.

Numerous population based algorithms have been identified, the majority of which have been exhaustively reviewed\cite{Beheshti2013}; however the genetic algorithm, particle swarm optimization and the gravitational search algorithms are among the most widely used population based algorithms that is used to solve optimization problems. The algorithm investigated in this research is an adaptation of the GSA, and the results presented in subsequent sections compares the quality of the Infrasonic Search Algorithm with the Genetic Algorithm, Particle Swarm Optimization Algorithm and the Gravitational Search Algorithm.

\section{Methods}

\subsection{Infrasonic Search Algorithm (ISA)}
Largely inspired from the Gravitational Search Algorithm (GSA),  in the Infrasonic Search Algorithm (ISA) agents aim to creation the loudest sound intensity by listening and adjusting its position based on the sound produced from other agents in the solution space.

Peafowls have been shown to produce infrasonic sounds with their tails to attract mates. Other peacocks respond to this mating sound by increased movement or an increase in a corresponding infrasonic sound.\cite{Freeman2015}  Adapting this for a heuristic search algorithm, high intensity sounds are created from high quality solutions which determine  the rate of movement/exploration of search space from neighbouring agents. Given that sound intensity reduces as the distance between the originating source increases, the rate at which an agent responds to sound from another agent reduces as the euclidean distance 
between the two agents increases. The formulation for updating and creating sounds in described in the next section.

\subsubsection{Formulation of the ISA}
The intensity of sound from a peafowl and the rate at which the sound is produced is used to attract mates among peafowls. This study uses this principle for a meta-heuristic search algorithm. In adapting the peafowls' infrasonic mating sound display for a meta-heuristic search algorithm, higher sound intensity/rate are created by fitter agents, and thus result in increased movement/exploration from other agents. 

Similar to the GSA, in the ISA agents are attracted towards other fitter agents who ostensibly are better mates, and the extent to which an agent will move towards another agent is based on the sound intensity from that agent and the distance between the two agents. 

The sound intensity between two agents represented as vectors $V_i$ and $V_j$ is defined as:
\begin{equation}
I_{i,j} = \frac{P_i}{A^2_{i,j}}
\label{eq1}
\end{equation}
where $A_{i,j} = 4 \pi r^2_{i,j}$ and $r$ is he euclidean distance between $V_i$ and $V_j$ and $P_i$ is the normalised power determined by the fitness of $V_i$.

Sound power is the rate at which the energy from a source (agent) is transferred. The normalised fitness at a given time $t$ of the agent is determined by Equation \ref{eq2} and Equation \ref{eq3}. 

\begin{equation}
p_i(t) = \frac{fitness_i(t) - worst(t)}{best(t) - worst(t)}
\label{eq2}
\end{equation}

\begin{equation}
P_i(t) = \frac{p_i(t)}{ \sum\limits_{j=1}^{n} p_j(t) }
\label{eq3}
\end{equation}

An agent moves towards its neighbour with the highest intensity by a percentage of the difference between the vector that represents the agent $V_i$ and the vector that represents the neighbour $V_j$ with the highest intensity as shown in Equation \ref{eq5}. The percentage $per_i(t)$ is computed by Equation \ref{eq4}, where $\rho = [0, 100]$ is a variable that controls the extent to which an agent is attracted to the neighbour with the highest intensity.  

\begin{equation}
per_i(t) =\begin{cases} \left(\frac{fitness_j(t)}{max(fitness_i(t), fitness_j(t))} ~ + ~ rand(0, 1) \right) \times \rho & minimisation \\
                     \left(\frac{fitness_i(t)}{max(fitness_i(t), fitness_j(t))} ~ + ~ rand(0, 1) \right) \times \rho &  maximisation
       \end{cases}
\label{eq4}
\end{equation}

\begin{equation}
displacement_i(t) = \frac{\left( V_i(t) - V_j(t) \right) \times per_i(t)}{100}
\label{eq5}
\end{equation}

The displacement computed in Equation \ref{eq5} is added to the vector representing the agent $V_i$ plus a fraction of its past movement to form a new vector for the subsequent iteration as shown in Equation \ref{eq7}. The idea is that an agent's memory also has an influence on its future movement, a concept also used in the Gravitational Search Algorithm.

\begin{equation}
\begin{array}{c}
movement_i(t) = movement_i(t - 1) \times rand(0, 1) + displacement_i(t) \\
where~movement_i(0)~=~0
\end{array}
\label{eq6}
\end{equation}

\begin{equation}
V_i(t + 1) = movement_i(t)  + V_i(t)
\label{eq7}
\end{equation}

In cases when updating an agent's vector result in a vector outside the search domain, values outside the domain are set to a random value in the search domain. 

Agents in a population are initialized to random values, for example $F_1(x)$ for \ref{tab2} an agent will be initialized to a random vector $[d_1, d_2 .... d_{n=50}]$; where $d_1$ is a value in range $[-100, 100]$.  In each iteration, the position of agents are adjusted based on equation \ref{eq5}, equation \ref{eq6} and equation \ref{eq7} and then the global best agent is returned in the final iteration. In terms of the time complexity of the ISA, as with the GSA, all agents are compared with others, thus the algorithm is run in $O(n^2)$. The pseudocode for the ISA is shown in Algorithm \ref{algo1}.

Two parameters control how the agents are moved towards neighbours; $\rho$ a value in range $[0 - 100]$ is a percentage that controls the extent to which an agent moves towards its highest intensity neighbour and  the random value in equation \ref{eq6} controls the extent to which an agents' historical movement influences its new position. Grid search is used to find appropriate values for $\rho$ for each benchmark function. In grid parameters are exhaustively searched to find the most optimal value for a particular algorithm.

\begin{algorithm2e}[H]
\DontPrintSemicolon
  
  $pop \gets initialize~population~to~random~values$\\
  $movement \gets initialize~historical~movement~to~zeroes$\\
  $best \gets fittest~solution~in~pop$
  
  \For{i = 1 to max iterations}    
        { 
        Compute displacement (Equation \ref{eq5})\\
        	Compute the extent to which agents are moved (Equation \ref{eq6})\\
        	Update historical movement\\
        	Update the position of agents to form a new population (Equation \ref{eq7})\\
        	Update best solution\\
  }
  
  \Return $best$.
\caption{Pseudocode for the Infrasonic Search Algorithm (ISA)}
\label{algo1}
\end{algorithm2e}

\section{Comparative Study}

Hussain et al.\cite{hussain2017common} have reviewed benchmark functions that can be used to evaluate the quality of meta-heuristic algorithms. 23 of these benchmark functions shown in  Appendix \ref{appendix_a} were used to evaluate to performance of the ISA in order to compare with the results of the Genetic Algorithm, Particle Swarm Optimization and the Gravitational Search Algorithm published by Rashedi et al.\cite{Rashedi2009}. 

Table \ref{tab2} shows the average and median results for the minimization of the 23 benchmark functions in Appendix \ref{appendix_a}  after 1000 runs for $F_1$ to $F_{13}$ and 500 iterations for $F_{14}$ to $F_{23}$. The result for the Genetic Algorithm, Particle Swarm Optimization and the Gravitational Search Algorithm is extracted from the results presented by Rashedi et al.\cite{Rashedi2009}.

\section{Discussion and Conclusion}
Simplistic methods have been shown to be unable to find adequate results for the majority
of optimisation problems, which is why this researches have focused on finding more suitable
techniques to find optimal solutions to these kind of problems.

Even as a number of search algorithms inspired from natural phenomena and/or other principles
that perform well for optimisation problems have been developed, the no free lunch theorem 
suggests that no method is guaranteed to outperform all others in all problems. Thus, there is the 
constant need to adapt previously developed algorithms or identify new algorithms that may perform 
well for some categories of search problems.

This research presented an algorithm inspired from the mating behaviour of peafowls. Peafowls create 
infrasonic sounds to attract possible mates, and the probability that another peafowl responds to to a 
generated infrasonic sound is increases based on the sound intensity produced and the distance between 
peafowls. This is a simplistic description on how peafowls identify suitable mates. In reality, infrasonic 
infrasonic sound generation is just one factor in a more complex process.

Following the principle of peafowls using infrasonic sounds to attract suitable mates, an agent in a population 
will produce sounds with higher intensity based on the quality (fitness) of the agents' solution, and thus attract 
other agents to itself. This idea is identical to the Gravitational Search Algorithm, in that in the GA, fitter 
solutions have higher masses and thus attracts other solutions. The main difference between the GSA 
and the method presented in this research is in the formulation in which agent move towards each other. In the GA, 
the formulation is based on the laws of gravity while the method presented in this study (Infrasonic Sound Algorithm), 
is based on the formulations of sound travelling in a medium.

The results showed in Table \ref{tab2} suggest that the ISA was able to identify competitive solutions to 23 benchmark unimodal
and multimodal functions compared to the Genetic Algorithm, Particle Swarm Optimization and the Gravitational Search Algorithm. The ISA 
performed better than the GSA in 4 of the 23 benchmark functions, and better than the Genetic Algorithm in 10 of the 23 
benchmark functions, suggesting that the proposed method may be suitable for some kind of optimization problems. Further work 
will aim to find those optimization problems that the ISA will be suitable for.

\bibliographystyle{unsrt}

\section{Appendices}
\begin{landscape}

\label{appendix_a}

\begin{table}[h]
\centering
\tiny
\caption{Evaluation results for the benchmark functions with the GA, PSA, GSA and ISA algorithms}{ 
\begin{tabular}{lccccccccc}
\hline
   & \multicolumn{2}{c}{$\scriptscriptstyle{GA}$} & \multicolumn{2}{|c}{$\scriptscriptstyle{PSA}$} & \multicolumn{2}{|c}{$\scriptscriptstyle{GSA}$} & \multicolumn{2}{|c}{$\scriptscriptstyle{ISA}$} \\
   & $\scriptscriptstyle{Average}$   & $\scriptscriptstyle{Median}$   & \multicolumn{1}{|c}{$\scriptscriptstyle{Average}$}    & $\scriptscriptstyle{Median}$    & \multicolumn{1}{|c}{$\scriptscriptstyle{Average}$}    & $\scriptscriptstyle{Median}$     & \multicolumn{1}{|c}{$\scriptscriptstyle{Average}$}    & $\scriptscriptstyle{Median}$    \\ \hline
$\scriptscriptstyle{F1}$ & $\scriptscriptstyle{23.13}$     & $\scriptscriptstyle{21.87}$     & \multicolumn{1}{|c}{$\scriptscriptstyle{1.8 \times 10 ^{-3}}$}   & $\scriptscriptstyle{5.0 \times 10 ^{-2}}$     & \multicolumn{1}{|c}{$\scriptscriptstyle{7.3 \times 10 ^{-11}}$}      & $\scriptscriptstyle{7.1 \times 10 ^{-11}}$          & \multicolumn{1}{|c}{$\scriptscriptstyle{14.93}$ }      & $\scriptscriptstyle{15.94}$       \\ 

$\scriptscriptstyle{F2}$ & $\scriptscriptstyle{1.07}$     & $\scriptscriptstyle{1.13}$    & \multicolumn{1}{|c}{$\scriptscriptstyle{2.0}$}      & $\scriptscriptstyle{1.9 \times 10^{-3}}$     & \multicolumn{1}{|c}{$\scriptscriptstyle{4.03 \times 10^{-5}}$ }      & $\scriptscriptstyle{4.07 \times 10^{-5}}$       & \multicolumn{1}{|c}{$\scriptscriptstyle{2.55}$ }      & $\scriptscriptstyle{2.62}$      \\ 

$\scriptscriptstyle{F3}$ & $\scriptscriptstyle{5.6 \times 10^{3}}$     & $\scriptscriptstyle{5.6 \times 10^{3}}$    & \multicolumn{1}{|c}{$\scriptscriptstyle{4.1 \times 10^{3}}$}      & $\scriptscriptstyle{2.2 \times 10^{3}}$     & \multicolumn{1}{|c}{$\scriptscriptstyle{0.16 \times 10^{3}}$ }      & $\scriptscriptstyle{0.15 \times 10^{3}}$       & \multicolumn{1}{|c}{$\scriptscriptstyle{0.99 \times 10^{3}}$}      & $\scriptscriptstyle{1.03 \times 10^{3}}$   \\

$\scriptscriptstyle{F4}$ & $\scriptscriptstyle{11.78}$     & $\scriptscriptstyle{11.94}$    & \multicolumn{1}{|c}{$\scriptscriptstyle{8.1}$}      & $\scriptscriptstyle{7.4}$     & \multicolumn{1}{|c}{$\scriptscriptstyle{3.7 \times 10^{-6}}$ }      & $\scriptscriptstyle{3.7 \times 10^{-6}}$       & \multicolumn{1}{|c}{$\scriptscriptstyle{5.2}$}           & $\scriptscriptstyle{5.8}$     \\ 

$\scriptscriptstyle{F5}$ & $\scriptscriptstyle{1.1 \times 10^3}$     & $\scriptscriptstyle{1.0 \times 10^3}$    & \multicolumn{1}{|c}{$\scriptscriptstyle{3.6 \times 10^4}$}      & $\scriptscriptstyle{1.7 \times 10^3}$     & \multicolumn{1}{|c}{$\scriptscriptstyle{25.16}$ }      & $\scriptscriptstyle{25.18}$       & \multicolumn{1}{|c}{$\scriptscriptstyle{2.31 \times 10^2}$}       & $\scriptscriptstyle{2.94 \times 10^{2}}$     \\ 

$\scriptscriptstyle{F6}$ & $\scriptscriptstyle{24.01}$     & $\scriptscriptstyle{24.55}$    & \multicolumn{1}{|c}{$\scriptscriptstyle{1.0 \times 10^{-3}}$}      & $\scriptscriptstyle{6.6 \times 10^{-3}}$     & \multicolumn{1}{|c}{$\scriptscriptstyle{8.3 \times 10^{-11}}$ }      & $\scriptscriptstyle{7.7 \times 10^{-11}}$       & \multicolumn{1}{|c}{$\scriptscriptstyle{15.53}$}      & $\scriptscriptstyle{16.29}$     \\

$\scriptscriptstyle{F7}$ & $\scriptscriptstyle{0.06}$     & $\scriptscriptstyle{0.06}$    & \multicolumn{1}{|c}{$\scriptscriptstyle{0.04}$}      & $\scriptscriptstyle{0.04}$     & \multicolumn{1}{|c}{$\scriptscriptstyle{0.018}$ }      & $\scriptscriptstyle{0.015}$       &  \multicolumn{1}{|c}{$\scriptscriptstyle{0.08}$ }      & $\scriptscriptstyle{0.08}$    \\

$\scriptscriptstyle{F8}$ & $\scriptscriptstyle{-1.2 \times 10^4}$     & $\scriptscriptstyle{-1.2 \times 10^4}$    & \multicolumn{1}{|c}{$\scriptscriptstyle{-9.8 \times 10^3}$}      & $\scriptscriptstyle{-9.8 \times 10^3}$     & \multicolumn{1}{|c}{$\scriptscriptstyle{-2.8 \times 10^3}$ }      & $\scriptscriptstyle{-2.6 \times 10^3}$       & \multicolumn{1}{|c}{$\scriptscriptstyle{1.09 \times 10^4}$ }      & $\scriptscriptstyle{1.09 \times 10^4}$    \\

$\scriptscriptstyle{F9}$ & $\scriptscriptstyle{5.90}$     & $\scriptscriptstyle{5.71}$    & \multicolumn{1}{|c}{$\scriptscriptstyle{55.1}$}      & $\scriptscriptstyle{56.6}$     & \multicolumn{1}{|c}{$\scriptscriptstyle{15.32}$ }      & $\scriptscriptstyle{14.42}$       & \multicolumn{1}{|c}{$\scriptscriptstyle{147.51}$ }      & $\scriptscriptstyle{149.25}$    \\

$\scriptscriptstyle{F10}$ & $\scriptscriptstyle{2.13}$     & $\scriptscriptstyle{2.16}$    & \multicolumn{1}{|c}{$\scriptscriptstyle{9.0 \times 10^{-3}}$}      & $\scriptscriptstyle{6.0 \times 10^{-3}}$     & \multicolumn{1}{|c}{$\scriptscriptstyle{6.9 \times 10^{-3}}$ }      & $\scriptscriptstyle{6.9 \times 10^{-3}}$       & \multicolumn{1}{|c}{$\scriptscriptstyle{3.78 }$ }      & $\scriptscriptstyle{3.81}$   \\

$\scriptscriptstyle{F11}$ & $\scriptscriptstyle{1.16}$     & $\scriptscriptstyle{1.14}$    & \multicolumn{1}{|c}{$\scriptscriptstyle{0.01}$}      & $\scriptscriptstyle{0.0081}$     & \multicolumn{1}{|c}{$\scriptscriptstyle{0.29}$ }      & $\scriptscriptstyle{0.04}$       & \multicolumn{1}{|c}{$\scriptscriptstyle{2.47}$ }      & $\scriptscriptstyle{2.52}$    \\

$\scriptscriptstyle{F12}$ & $\scriptscriptstyle{0.051}$     & $\scriptscriptstyle{0.039}$    & \multicolumn{1}{|c}{$\scriptscriptstyle{0.29}$}      & $\scriptscriptstyle{0.11}$     & \multicolumn{1}{|c}{$\scriptscriptstyle{0.01}$ }      & $\scriptscriptstyle{4.2 \times 10 ^{-13}}$       & \multicolumn{1}{|c}{$\scriptscriptstyle{4.18}$}      & $\scriptscriptstyle{5.09}$    \\ 

$\scriptscriptstyle{F13}$ & $\scriptscriptstyle{0.081}$     & $\scriptscriptstyle{0.032}$    & \multicolumn{1}{|c}{$\scriptscriptstyle{3.1 \times 10^{-18}}$}      & $\scriptscriptstyle{2.2 \times 10^{-23}}$     & \multicolumn{1}{|c}{$\scriptscriptstyle{3.2 \times 10^{-32}}$ }      & $\scriptscriptstyle{2.3 \times 10^{-32}}$       & \multicolumn{1}{|c}{$\scriptscriptstyle{1.28}$ }      & $\scriptscriptstyle{1.33}$     \\ 

$\scriptscriptstyle{F14}$ & $\scriptscriptstyle{0.998}$     & $\scriptscriptstyle{0.998}$    & \multicolumn{1}{|c}{$\scriptscriptstyle{0.998}$}      & $\scriptscriptstyle{0.998}$     & \multicolumn{1}{|c}{$\scriptscriptstyle{3.70}$ }      & $\scriptscriptstyle{2.07}$       & \multicolumn{1}{|c}{$\scriptscriptstyle{0.998}$}      & $\scriptscriptstyle{0.998}$     \\

$\scriptscriptstyle{F15}$ & $\scriptscriptstyle{4.0 \times 10^{-3}}$     & $\scriptscriptstyle{1.7 \times 10^{-3}}$    & \multicolumn{1}{|c}{$\scriptscriptstyle{2.8 \times 10^{-3}}$}      & $\scriptscriptstyle{7.1 \times 10^{-4}}$     & \multicolumn{1}{|c}{$\scriptscriptstyle{8.0 \times 10^{-3}}$ }      & $\scriptscriptstyle{7.4 \times 10^{-4}}$       & \multicolumn{1}{|c}{$\scriptscriptstyle{8.7 \times 10^{-4}}$ }      & $\scriptscriptstyle{7.5 \times 10^{-4}}$ \\  

$\scriptscriptstyle{F16}$ & $\scriptscriptstyle{-1.0313}$     & $\scriptscriptstyle{-1.0315}$    & \multicolumn{1}{|c}{$\scriptscriptstyle{-1.0316}$}      & $\scriptscriptstyle{-1.0316}$     & \multicolumn{1}{|c}{$\scriptscriptstyle{-1.0316}$ }      & $\scriptscriptstyle{-1.0316}$       & \multicolumn{1}{|c}{$\scriptscriptstyle{-1.0186}$ }      & $\scriptscriptstyle{-1.0316}$   \\

$\scriptscriptstyle{F17}$ & $\scriptscriptstyle{0.3996}$     & $\scriptscriptstyle{0.3980}$    & \multicolumn{1}{|c}{$\scriptscriptstyle{0.3979}$}      & $\scriptscriptstyle{0.3979}$     & \multicolumn{1}{|c}{$\scriptscriptstyle{0.3979}$ }      & $\scriptscriptstyle{0.3979}$       & \multicolumn{1}{|c}{$\scriptscriptstyle{0.3979}$ }      & $\scriptscriptstyle{0.3979}$   \\

$\scriptscriptstyle{F18}$ & $\scriptscriptstyle{5.70}$     & $\scriptscriptstyle{3.0}$    & \multicolumn{1}{|c}{$\scriptscriptstyle{3.0}$}      & $\scriptscriptstyle{3.0}$     & \multicolumn{1}{|c}{$\scriptscriptstyle{3.0}$ }      & $\scriptscriptstyle{3.0}$       & \multicolumn{1}{|c}{$\scriptscriptstyle{3.0}$ }      & $\scriptscriptstyle{3.0}$    \\

$\scriptscriptstyle{F19}$ & $\scriptscriptstyle{-3.8627}$     & $\scriptscriptstyle{-3.8628}$    & \multicolumn{1}{|c}{$\scriptscriptstyle{-3.8628}$}      & $\scriptscriptstyle{-3.8628}$     & \multicolumn{1}{|c}{$\scriptscriptstyle{-3.8628}$ }      & $\scriptscriptstyle{-3.8628}$       &\multicolumn{1}{|c}{$\scriptscriptstyle{-3.8623}$ }      & $\scriptscriptstyle{-3.8625}$  \\

$\scriptscriptstyle{F20}$ & $\scriptscriptstyle{-3.3099}$     & $\scriptscriptstyle{-3.3217}$    & \multicolumn{1}{|c}{$\scriptscriptstyle{-3.2369}$}      & $\scriptscriptstyle{-3.2031}$     & \multicolumn{1}{|c}{$\scriptscriptstyle{-2.0569}$ }      & $\scriptscriptstyle{-1.9946}$       & \multicolumn{1}{|c}{$\scriptscriptstyle{-3.2829}$ }      & $\scriptscriptstyle{-3.3188}$     \\

$\scriptscriptstyle{F21}$ & $\scriptscriptstyle{-5.6605}$     & $\scriptscriptstyle{-2.6824}$    & \multicolumn{1}{|c}{$\scriptscriptstyle{-6.6290}$}      & $\scriptscriptstyle{-5.1008}$     & \multicolumn{1}{|c}{$\scriptscriptstyle{-6.0748}$ }      & $\scriptscriptstyle{-5.0552}$       & \multicolumn{1}{|c}{$\scriptscriptstyle{-9.6767}$ }      & $\scriptscriptstyle{-10.0616}$    \\

$\scriptscriptstyle{F22}$ & $\scriptscriptstyle{-7.3421}$     & $\scriptscriptstyle{-10.3932}$    & \multicolumn{1}{|c}{$\scriptscriptstyle{-9.1118}$}      & $\scriptscriptstyle{-10.402}$     & \multicolumn{1}{|c}{$\scriptscriptstyle{-9.3399}$ }      & $\scriptscriptstyle{-10.402}$       & \multicolumn{1}{|c}{$\scriptscriptstyle{-9.9791}$ }      & $\scriptscriptstyle{-10.3863}$    \\

$\scriptscriptstyle{F23}$ & $\scriptscriptstyle{-6.2541}$     & $\scriptscriptstyle{-4.5054}$    & \multicolumn{1}{|c}{$\scriptscriptstyle{-9.7634}$}      & $\scriptscriptstyle{-10.536}$     & \multicolumn{1}{|c}{$\scriptscriptstyle{-9.4548}$ }      & $\scriptscriptstyle{-10.536}$       & \multicolumn{1}{|c}{$\scriptscriptstyle{-9.9059}$ }      & $\scriptscriptstyle{-10.4692}$     \\

\hline
\end{tabular}}
\label{tab2}
\end{table}

\begin{table}[h!]

\centering
\tiny
{
\caption{23 unimodal and multimodal benchmark functions for evaluating the efficiency of the Infrasonic Search Algorithm.}
\begin{tabular}{lllccc}
\hline
   Function   & Equation   & Domain &   Dim(n) &  Pop. size    & iterations    \\ 
\hline
   $\scriptscriptstyle{F_1(X)}$   &  $ \scriptscriptstyle{ \sum \limits _{i=1}^{n} x_i^2 }$         & [-100, 100] & 30      &    50           & 1000       \\[0.1cm]
   $\scriptscriptstyle{F_2(X)}$   &  $\scriptscriptstyle{ \sum \limits _{i=1}^{n} \left\lvert x_i \right\rvert +  \prod \limits_{i=1}^n \left\lvert x_i \right\lvert }$                    & [-10, 10] & 30      &    50           & 1000            \\[0.1cm]
   $\scriptscriptstyle{F_3(X)}$   & $\scriptscriptstyle{ \sum \limits_ {i=1}^{n} \left( \sum \limits_{j=1}^i x_j \right)^2 }$  & [-100, 100] & 30 &  50  & 1000  \\ [0.1cm]
   $\scriptscriptstyle{F_4(X)}$   &  $\scriptscriptstyle{ _i ^{max} \left\lbrace \left \lvert x_i \right \rvert, 1 \leqslant i \leqslant n \right \rbrace }$  & [-100, 100] & 30 &  50  & 1000  \\[0.1cm]
      $\scriptscriptstyle{F_5(X)}$   & $\scriptscriptstyle{  \sum \limits _{i =1} ^{n-1} \left[ 100 \left( x_{i +1} - x_i^2 \right)^2 + \left( x_i - 1 \right)^2 \right] }$  & [-30, 30] & 30 &  50  & 1000  \\[0.1cm]
            $\scriptscriptstyle{F_6(X)}$   &  $\scriptscriptstyle{ \sum \limits _{i =1} ^{n} \left( \left[ x_i + 0.5 \right]\right) ^ 2 }$  & [-100, 100] & 30 &  50  & 1000  \\[0.1cm]
$\scriptscriptstyle{F_7(X)}$   &  $ \scriptscriptstyle{  \sum \limits _{i =1} ^{n} ix_i^4 + random\left[ 0, 1 \right) }$  & [-1.28, 1.28] & 30 &  50  & 1000  \\[0.1cm]
$\scriptscriptstyle{F_8(X)}$   &  $ \scriptscriptstyle{ \sum \limits _{i =1} ^{n} -x_i sin \left( \sqrt{ \left|x_i \right| } \right) }$  & [-500, 500] & 30 &  50  & 1000  \\[0.1cm]
$\scriptscriptstyle{F_9(X)}$   &  $\scriptscriptstyle{ \sum \limits _{i =1} ^{n} \left[ x_i^2 - 10 cos \left( 2 \pi x_i \right) + 10  \right] }$  & [-5.12, 5.12] & 30 &  50  & 1000  \\[0.1cm]
$\scriptscriptstyle{F_{10}(X)}$   &  $\scriptscriptstyle{ -20 exp \left(-0.2\sqrt{\frac{1}{n} \sum \limits _{i =1}^n x_i^2}\right) - exp \left(\frac{1}{n} \sum \limits _{i =1}^n cos \left( 2 \pi x_i \right)\right) + 20 + e }$  & [-32, 32] & 30 &  50  & 1000  \\[0.1cm]
$\scriptscriptstyle{F_{11}(X)}$   &  $ \scriptscriptstyle{ \frac{1}{4000} \sum \limits _{i = 1}^n x_i^2 - \prod \limits _{i =1} ^ n cos \left( \frac{x_i}{\sqrt{i}} \right) + 1 }$  & [-600, 600] & 30 &  50  & 1000  \\[0.1cm]
$\scriptscriptstyle{F_{12}(X)}$   & $ \scriptscriptstyle{ \lbrace \frac{x}{n}  10 sin\left( \pi y_1 \right) + \sum \limits _{i=1}^{n-1} \left( y_i -1 \right)^2 \left[1 + 10 sin^2 \left( \pi y_{i + 1} \right) \right] + \left( y_n -1 \right)^2 \rbrace   }$ & [-50, 50] & 30 &  50  & 1000  \\
& $ \scriptscriptstyle{ + \sum \limits _{i=1}^n  u\left( x_i, 10, 100, 4 \right) } $ &  &  &    &   \\[0.1cm]
 & $ \scriptscriptstyle{ y_i = 1 + \frac{x_i + 1}{4} } $ &  &  &    &   \\
  & $ \scriptscriptstyle{ u(x_i, a, k, m)} = \left\{\begin{smallmatrix}
 k(x_i - a)^m \hfill & x_i > a \hfill \\ 
 0 \hfill & -a < x_i < a \hfill \\ 
 k(-x_i - a)^m \hfill  & x_i < -a \hfill
\end{smallmatrix} \right. $ &  &  &    &   \\[0.1cm]
$\scriptscriptstyle{F_{13}(X)}$   &  $\scriptscriptstyle{ 0.1\lbrace sin^2(3 \pi x_1) + \sum \limits _{i =1} ^ n (x_i - 1)^2 [ 1 + sin^2(3 \pi x_i + 1)] + (x_n - 1)^2[1 + sin^2(2 \pi x_n)]  \rbrace} $  & [-50, 50] & 30 &  50  & 1000  \\
  &  $ \scriptscriptstyle{  + \sum \limits _{i = 1}^n u(x_i, 5, 100, 4) }$  &  &  &    &   \\[0.4cm]
$\scriptscriptstyle{F_{14}(X)}$   &  $  \scriptscriptstyle{ \left( \frac{1}{500} + \sum \limits _{j =1} ^{25} \frac{1}{j +  \sum \limits _{i = 1} ^ 2 (x_i - a_{ij})^6} \right)^ {-1} }$  & [-65.53, 65.53] & 2 &  50  & 500  \\[0.1cm]
$\scriptscriptstyle{F_{15}(X)}$   &  $\scriptscriptstyle{ \sum \limits _{i =1} ^{11} \left[ a_i - \frac{x_1(b_i^2 + b_ix_2)}{b_i^2 + b_ix_3 + x_4} \right] ^ 2} $  & [-5, 5] & 4 &  50  & 500  \\[0.1cm]
$\scriptscriptstyle{F_{16}(X)}$   &  $ \scriptscriptstyle{ 4x_1^2 - 2.1x_1^4 + \frac{1}{3}x_1^6 + x_1x_2 - 4x_2^2 +4x_2^4}$  & [-5, 5] & 2 &  50  & 1000  \\[0.1cm]
$\scriptscriptstyle{F_{17}(X)}$   &  $\scriptscriptstyle{ \left( x_2 + \frac{5.1}{4 \pi ^2} x_1^2 + \frac{5}{\pi} x_1 - 6 \right) ^2 + 10\left(1 - \frac{1}{8 \pi} \right) cosx_1 + 10 }$  & [-5, 10] x [0, 15] & 2 &  50  & 500  \\[0.1cm]
$\scriptscriptstyle{F_{18}(X)}$   &  $\scriptscriptstyle{ [1 + (x_1 + x_2 + 1)^2(19 -14x_1 + 3x_1^2 - 14x_2 + 6x_1x_2 +3x_2^2)] }$  & [-5, 5] & 2 &  50  & 500  \\
   &  $ \scriptscriptstyle{ \times [30 +  (2x_1 - 3x_2)^2 \times (18 - 32x_1 + 12x_1^2 + 48x_2 - 36x_1x_2 + 27x_2^2)]}$  &  &  &    &   \\[0.1cm]
$\scriptscriptstyle{F_{19}(X)}$   &  $\scriptscriptstyle{ - \sum \limits _{i=1} ^4 c_i exp \left(- \sum \limits _{j=1} ^3 a_{ij} (x_j - p_{ij})^2 \right) }$  & [0, 1] & 3 &  50  & 500  \\[0.1cm]
$\scriptscriptstyle{F_{20}(X)}$   &  $\scriptscriptstyle{ - \sum \limits _{i=1} ^4 c_i exp \left(- \sum \limits _{j=1} ^6 a_{ij} (x_j - p_{ij})^2 \right) }$  & [0, 1] & 6 &  50  & 500  \\[0.1cm]
$\scriptscriptstyle{F_{21}(X)}$   &  $\scriptscriptstyle{ - \sum \limits _{i = 1}^ 5 [(X - a_i)(X - a_i)^T + c_i] ^ {-1} }$  & [0, 10] & 4 &  50  & 500  \\[0.1cm]
$\scriptscriptstyle{F_{22}(X)}$   &  $\scriptscriptstyle{ - \sum \limits _{i = 1}^ 7 [(X - a_i)(X - a_i)^T + c_i] ^ {-1} }$  & [0, 10] & 4 &  50  & 500  \\[0.1cm]
$\scriptscriptstyle{F_{23}(X)}$   &  $\scriptscriptstyle{ - \sum \limits _{i = 1}^ {10} [(X - a_i)(X - a_i)^T + c_i] ^ {-1} }$  & [0, 10] & 4 &  50  & 500  \\[0.1cm]
\hline
\end{tabular}}

\end{table}
\end{landscape}

\end{document}